\documentclass[letterpaper, 10 pt, conference]{ieeeconf}  % Comment this line out if you need a4paper

\IEEEoverridecommandlockouts                              % This command is only needed if 
\usepackage{graphics} % for pdf, bitmapped graphics files
\usepackage{epsfig} % for postscript graphics files
\usepackage{amsmath} % assumes amsmath package installed
\usepackage{amssymb}  % assumes amsmath package installed
\usepackage{cite}

\usepackage{algorithm}
\usepackage{algorithmic}

\usepackage{multirow}
\usepackage{booktabs}
\usepackage{threeparttable}
\usepackage{bbold}
\usepackage{pifont}

\makeatletter
\let\MYcaption\@makecaption
\makeatother

\usepackage[font=footnotesize]{subcaption}

\makeatletter
\let\@makecaption\MYcaption
\makeatother

\usepackage{subcaption}
\def\ie{{\it i.e.}}
\def\etal{{\it et al.}}

\title{\LARGE \bf 
Deep Probabilistic Traversability with Test-time Adaptation for Uncertainty-aware Planetary Rover Navigation
}

\author{Masafumi Endo$^{1}$, Tatsunori Taniai$^{2}$, %Ryo Yonetani$^{2}$, 
and Genya Ishigami$^{1}$% <–this % stops the space
\thanks{%Manuscript received: Month Day, Year; Revised: Month Day, Year; Accepted: Month Day, Year. This paper was recommended for publication by the Editor [Name Here] upon evaluation by the Associate Editor and Reviewers’ comments.
This work was supported in part by JSPS KAKENHI Grant Number JP22J22731 and advised by Ryo Yonetani during its early stages. (\emph{Corresponding author: Masafumi Endo}). }
\thanks{$^{1}$Masafumi Endo and Genya Ishigami are with the Space Robotics Group, Department of Mechanical Engineering, Keio University Yokohama 223-8522, Japan
        (e-mail: masafumi.endo@keio.jp; ishigami@mech.keio.ac.jp).
        }
\thanks{$^{2}$Tatsunori Taniai is with OMRON SINIC X Corporation, Tokyo 113-0033, Japan 
        (e-mail: tatsunori.taniai@sinicx.com).
        }%
}

\usepackage{color}

\usepackage{textcomp}
\usepackage{fancyhdr}

\begin{document}

\twocolumn[
\noindent
© 2024 IEEE. Personal use of this material is permitted. Permission from IEEE must be obtained for all other uses, in any current or future media, including reprinting/republishing this material for advertising or promotional purposes, creating new collective works, for resale or redistribution to servers or lists, or reuse of any copyrighted component of this work in other works.\\

\noindent
\textbf{Submitted article:}\\
M. Endo, T. Taniai, and G. Ishigami ``Deep Probabilistic Traversability with Test-time Adaptation for Uncertainty-aware Planetary Rover Navigation,'' \textit{Under review for IEEE Robotics and Automation Letters}, 2024.
]
\thispagestyle{empty}
\pagenumbering{gobble}
\clearpage

\maketitle
\thispagestyle{empty}
\pagestyle{empty}

%%%%%%%%%%%%%%%%%%%%%%%%%%%%%%%%%%%%%%%%%%%%%%%%%%%%%%%%%%%%%%%%%%%%%%%%%%%%%%%%
\begin{abstract}

Traversability assessment of deformable terrain is vital for safe rover navigation on planetary surfaces. 
Machine learning (ML) is a powerful tool for traversability prediction but faces predictive uncertainty. This uncertainty leads to prediction errors, increasing the risk of wheel slips and immobilization for planetary rovers.
To address this issue, we integrate principal approaches to uncertainty handling---quantification, exploitation, and adaptation---into a single learning and planning framework for rover navigation.
The key concept is \emph{deep probabilistic traversability}, forming the basis of an end-to-end probabilistic ML model that predicts slip distributions directly from rover traverse observations.
This probabilistic model quantifies uncertainties in slip prediction and exploits them as traversability costs in path planning.
Its end-to-end nature also allows adaptation of pre-trained models with in-situ traverse experience to reduce uncertainties.
We perform extensive simulations in synthetic environments that pose representative uncertainties in planetary analog terrains.
Experimental results show that our method achieves more robust path planning under novel environmental conditions than existing approaches.

\end{abstract}

%%%%%%%%%%%%%%%%%%%%%%%%%%%%%%%%%%%%%%%%%%%%%%%%%%%%%%%%%%%%%%%%%%%%%%%%%%%%%%%%
\section{Introduction}

Autonomy is crucial for rapid and extensive robotic planetary exploration due to significant communication delays between Earth and space.  
In NASA's Mars 2020 mission, the Perseverance rover's autonomous navigation system (AutoNav~\cite{goldberg2002stereo}) averaged a traverse distance of 144.4 meters per Martian solar day (24 h 39 min~\cite{hecht2009detection}). 
Although this performance surpasses that of prior rovers~\cite{verma2023autonomous}, AutoNav has room for improvement in challenging environments, as it primarily achieved rapid traverse on benign terrain.
While AutoNav evaluates \textit{traversability} by detecting geometric obstacles on Martian terrain~\cite{goldberg2002stereo}, deformable surfaces covered with fine-grained regolith are more hazardous for rovers than apparent obstacles.
These deformable surfaces induce excessive wheel slip, degrade driving speed, increase energy consumption, and eventually cause permanent entrapment in loose regolith. 
In one case, NASA's Opportunity rover spent more than six weeks trapped in the rippled sand of the Purgatory Dune, resulting in a prolonged mission timeline~\cite{squyres2006overview}. 
Hence, reliable traversability assessment of wheel slip is essential to ensure the safety of rovers traversing unexplored regions.

Traversability assessment of deformable terrain is challenging because of the complex mechanical interactions arising from rover mobility mechanisms and terrain characteristics, such as mechanical properties and surface geometry~\cite{wong2008theory}.
ML can be used to derive latent relationships between environmental features and slip behaviors from training samples.
However, despite advances in robot perception and planning~\cite{kober2013reinforcement,grigorescu2020survey}, no ML model can guarantee perfect predictions owing to inherent \emph{predictive uncertainty}. 
This uncertainty leads to erroneous predictions, potentially causing the unrecoverable immobilization of rovers on distant planets.
%Given this risk, 
Thus, addressing uncertainty in traversability prediction is essential for successful rover navigation.

In ML-based traversability prediction, uncertainty can be managed via three principal approaches.
\emph{Quantification} of uncertainty, usually through probabilistic models, provides the degree of uncertainty in the prediction.
\emph{Exploitation} of measured uncertainty in planning and decision-making processes helps mitigate the risks of potential prediction errors.
\emph{Adaptation} of pre-trained ML models during navigation is also effective in reducing uncertainty due to limited prior observations.
Regarding distant planets, ML-based navigation struggles with limited prior traverse experience and uncontrollable environmental factors, such as lighting conditions, which can cause \emph{domain shifts} between training and test environmental conditions.

\begin{figure*}[t]
    \centering
    \includegraphics[width=1.\hsize]{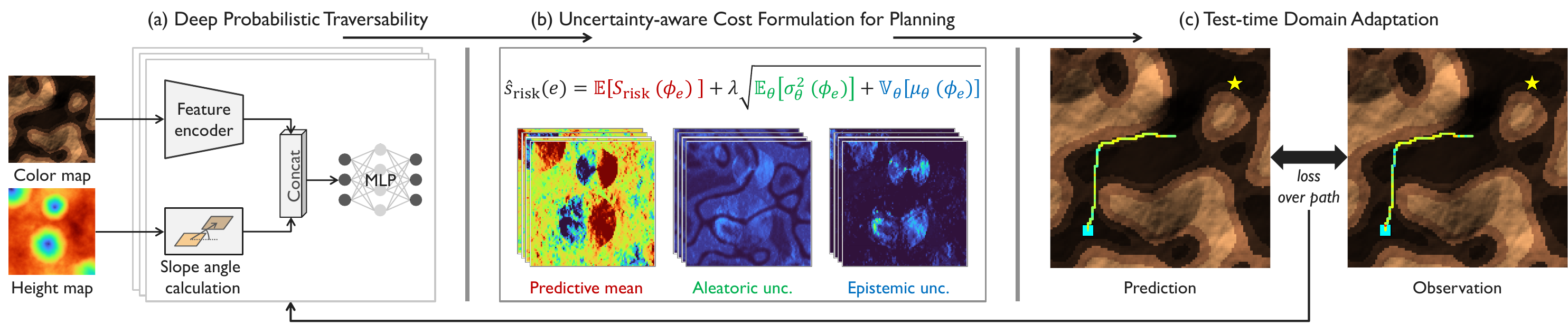}
    \caption{
Overview of the proposed unified framework for planetary rover navigation, consisting of (a) deep probabilistic traversability, (b) uncertainty-aware path planning, and (c) test-time DA with in-situ slip measurements, forming an iterative loop to address uncertain traversability prediction.
}
    \label{fig:overview}
\end{figure*}

In this study, we integrate these three uncertainty handling approaches, namely uncertainty quantification and exploitation and model adaptation, into a unified learning and planning framework for safe rover navigation.
The key concept in uniting these approaches is \emph{deep probabilistic traversability}, an end-to-end probabilistic ML model using deep neural networks (DNNs) for traversability prediction (Fig.~\ref{fig:overview}a).
This model predicts the rover's wheel slip directly from visual and geometric observations of the target environment while quantifying the predictive uncertainty via probability distributions.
Subsequently, this uncertainty can be exploited to reduce the risks associated with imperfect predictions in path planning (Fig.~\ref{fig:overview}b) and is further minimized by adapting the learned distributions to in-situ slip measurements during path execution (Fig.~\ref{fig:overview}c).

We extensively evaluate our method in synthetic celestial environments, focusing on the challenges posed by out-of-domain (OOD) geometry and appearance observations.
Comparisons with existing methods demonstrate that our method, with uncertainty-aware planning and domain adaptation (DA), achieves safer rover navigation under novel environmental conditions.

\section{Related Work}

ML-based traversability prediction is a key component of off-road autonomy, where \emph{visual} and \emph{geometric} cues are essential.
Visual cues capture the diverse geological semantic features that influence robot behavior.
Thus, classification methods use the semantic segmentation of appearance imagery to categorize symbolic terrain types and identify traversable areas~\cite{filitchkin2012feature,rothrock2016spoc}.
This appearance-based classification is often combined with geometry-based regression methods to provide further variation in traversability within each class. Classify-then-regress methods model the traversability for each terrain class by correlating the degree of wheel slip with the terrain inclination~\cite{cunningham2017locally,endo2021terrain,inotsume2021adaptive}. Mixture-of-experts (MoE) methods further account for terrain class likelihoods for slip regressors\cite{angelova2007learning,endo2023risk}. Although these combined approaches using visual and geometric cues are effective, they require separate training processes for the classifier and regressor models, which involve human supervision.

Using the measured traverse data for supervision, learning traversability from experience has become a viable alternative.
This approach provides a direct interpretation of robot--terrain interactions and eliminates the need for labor-intensive manual annotation.
Previous studies often used past driving trajectories to auto-generate ground-truth traversability, which assumes scene familiarity with robots~\cite{schmid2022self,seo2023learning,jung2023vstrong}.
Proprioceptive signals, such as force-torque~\cite{welhausen2019where} and linear acceleration~\cite{castro2023does}, define traversability in continuous-valued regression models.
The absence of manual annotation also facilitates model adaptation through incremental learning with in-situ measurements, improving the robustness to novel situations~\cite{chen2023learning,frey2023fast}.
Our method builds on this learning-from-experience paradigm with an end-to-end DNN backbone, while exploiting both visual and geometric cues.

Off-road navigation algorithms leverage traversability-prediction techniques to ensure robot safety in unstructured environments.
In planetary exploration, rover navigation systems adopt classification models to identify traversable regions~\cite{ono2015risk} and regression models to derive continuous traversability costs from wheel slip~\cite{helmick2009terrain}.
In terrestrial settings, some studies~\cite{gasparino2022wayfast,gasparino2024wayfaster,meng2023terrainnet} employed traversability models within the model predictive path integral (MPPI) control framework for off-road navigation. However, these methods do not exploit the uncertainty in traversability prediction for safer navigation, as they often rely on deterministic models~\cite{gasparino2022wayfast,gasparino2024wayfaster,meng2023terrainnet}.
By contrast, we introduce uncertainty quantification in the prediction models to estimate the reliability of predictions and reduce the risk of rover immobility.

Uncertainty quantification has been extensively studied in ML, including the Bayesian interpretation~\cite{williams2006gaussian,mackay1992practical}, Monte Carlo dropout~\cite{gal2016dropout}, and model ensembles~\cite{lakshminarayanan2017simple} (see \cite{gawlikowski2023survey} for details).
In robotic decision-making, uncertainty quantification offers valuable tools for risk assessment and is used as a constraint or cost.
For instance, chance-constraint formulations have been explored for risk-aware off-road navigation~\cite{inotsume2020robust,endo2022active,candela2022approach}; however, these constraint-based methods are better suited for Boolean events rather than a range of risks, such as slip.
Conversely, cost-based methods estimate conservative traversability by incorporating uncertainty into costs~\cite{takemura2023uncertainty, lee2023learning, barbosa2021risk, fan2021step, endo2023risk, feng2023learning}. 
This approach involves designing custom-uncertainty-aware traveling costs~\cite{takemura2023uncertainty, lee2023learning} or applying statistical risk assessments~\cite{cai2022risk,cai2022probabilistic,cai2023evora}, such as conditional value-at-risk (CVaR)~\cite{majumdar2020how}. %, which provides a pessimistic estimate of a random variable.
Other work explores reinforcement learning approaches that can implicitly exploit uncertainty to learn conservative navigation policies~\cite{feng2023learning}.
We employ cost-based uncertainty integration to express a range of risks while allowing for hard-obstacle constraints in regions where rover immobilization is likely to occur.

Alternative to uncertainty avoidance, DA reduces uncertainties and improves predictions by using new observations in test environments.
For planetary exploration, Hedrick~\etal~\cite{hedrick2020terrain} proposed an adaptive navigation system that updates a velocity map online using a Markov random field model.
For terrestrial navigation, adaptation using traverse experiences as self-supervision has been studied~\cite{chen2023learning,frey2023fast}. However, their deterministic predictions, which lack uncertainty modeling, may pose serious risks to planetary robots.
Kim~\etal~\cite{kim2023bridging} presented an active DA approach for probabilistic traversability but it assumes human monitoring during the adaptation phase before deployment. Our deep probabilistic traversability model adapts \emph{on the fly} to in-situ slip measurements during rover path execution while predicting the occurrence of uncertain and risky regions.

Overall, we focus on planetary rover navigation under the risk of immobility due to uncertain slip on deformable terrain.
Our findings quantify, exploit, and reduce the uncertainty in traversability prediction within a unified framework, addressing challenges that have only been partially addressed by existing methods~\cite{endo2023risk,lee2023learning,kim2023bridging,cai2022risk,cai2022probabilistic,cai2023evora}.

\section{Preliminaries}

This section outlines a general formulation for path planning and traversability in terms of wheel slip and presents a target map representation for predicting traversability.

\subsection{Path Planning}

The path planning objective is to determine a series of feasible state transitions that allow a robot to navigate toward its destination safely and efficiently. 
Consider a grid map $\mathcal{G}=(\mathcal{V},\mathcal{E})$ representing an environment in 2D space, where $\mathcal{V}$ is a finite set of nodes $v$ enumerating possible robot states, and $\mathcal{E}$ is a set of edges $e$ representing state transitions from each node to its neighbors.
We follow a widely used setting, where $\mathcal{G}$ denotes a four-neighbor grid environment with evenly divided grids.
Every edge has a strictly positive cost calculated by a travel cost function $f_{\text{cost}}(v,v')$. 
The optimal path planning problem is formulated to find a path $\mathcal{P}=\left\{v_1, v_2,..., v_{|\mathcal{P}|}\right\}$, from the start $v_1=v_{\text{start}}\in\mathcal{V}$ to the goal $v_{|\mathcal{P}|}=v_{\text{goal}}\in\mathcal{V}$, with the minimum possible total cost defined as follows:
\begin{equation}
\label{eq:path_planning}
\min_\mathcal{P} \sum_{i=1}^{|\mathcal{P}|-1}f_{\text{cost}}\left(v_i, v_{i+1}\right).
\end{equation}

\subsection{Traversability as Wheel Slip}

Wheeled rovers experience slipping when traversing a deformable terrain. 
We consider the longitudinal slip ratio $s$~\cite{wong2008theory} to quantify rover traversability as follows:
\begin{equation}
\label{eq:slip_ratio}
{s}=\begin{cases}
    \left({u}_{\mathrm{ref}} - {u}\right)/{u}_{\mathrm{ref}},&{u} \leq {u}_{\mathrm{ref}}: \text{driving state},\\
    \left({u}_{\mathrm{ref}} - {u}\right) / {u},&{u} > {u}_{\mathrm{ref}}: \text{braking state},
\end{cases}
\end{equation}
where ${u}$ and ${u}_{\mathrm{ref}}$ denote the actual and reference velocities in the longitudinal direction, respectively.
A positive slip $s \in [0, 1]$ indicates a traverse slower than the one commanded, with higher values increasing travel time and, ultimately, causing rover immobilization on deformable terrain.
Conversely, a large negative slip $s \in (-1, 0]$ indicates a traverse faster than the one commanded, which can lead to rover dyscontrol when $s \approx -1$.
Although a higher velocity may seem preferable for reducing travel time, it poses operational risks, including deviations from the planned path and potential tip-overs.

\subsection{Map Representation}

Unlike typical path-planning problems, such as shortest pathfinding in 2D maps, our formulation for rover traveling requires variable traversability costs to account for slip behavior on deformable terrain.
While the actual slip trend is unknown, it primarily depends on terrain geological characteristics, such as mechanical properties and surface geometry in the given environments.
Thus, we associate the environment graph with the appearance and geometry of the terrain, particularly terrain surface colors and 3D positions, to predict traversability using ML models.
Projecting this information onto graph vertices is feasible for exploration missions, as the HiRISE camera used in Mars exploration captures overhead imagery with up to 0.5 m resolution~\cite{mcewen2007mars}.

\section{Uncertainty-aware Navigation Algorithm}

This section presents a unified framework for rover navigation that quantifies, exploits, and reduces the uncertainties to enhance the safety of the navigation process. %, which assesses the risk of immobility in planning. 
The overall flow of the method is illustrated in Fig.~\ref{fig:overview} and summarized as follows:
Given the appearance and geometry information, we first use the deep probabilistic traversability (DPT) to predict slips and their uncertainties as slip distributions (Section \ref{section:dpt}, Fig.~\ref{fig:overview}a).
Subsequently, we convert these distributions into traversability costs, accounting for potential prediction errors (Section \ref{section: ucf}, Fig.~\ref{fig:overview}b). 
These costs enable path-planning searches using (\ref{eq:path_planning}) to determine a path that minimizes the risk of rover immobilization.
Finally, we extend this process by incorporating the test-time adaptation of DPT and replanning using in-situ slip observations (Section \ref{section:adaptation}, Fig.~\ref{fig:overview}c).

\subsection{Deep Probabilistic Traversability Model}
\label{section:dpt}
We model the DPT using an ensemble of $M$ probabilistic DNNs~\cite{lakshminarayanan2017simple}.
We assume an identical architecture for these DNNs with different parameters $\theta_m$ for each network, denoted as $f_{\theta_m}$.
Each network takes a bird’s-eye view environment map of $H \times W$ nodes as the input (Fig.~\ref{fig:overview}a, left). The map represents node-wise RGB colors $I_v \in \mathbb{R}^{3}$ as appearance cues and edge-wise terrain pitch angles $\phi_e \in \mathbb{R}$ associated with the rover’s four possible movements at each node as geometric cues. Each network predicts a latent slip distribution for each edge $e$ by predicting the mean and variance of the Gaussian distribution $\mathcal{N}(\mu, \sigma^2)$ as follows:
\begin{equation}
[\mu_{e,\theta_m}, \sigma^2_{e,\theta_m}] = f_{\theta_m}(I, \phi_e). \label{eq:func_f}
\end{equation}
The DPT then models the slip on each edge $e$ as a random variable $S_e$ following a mixture of Gaussian distributions as
\begin{equation}
S_e \stackrel{\Delta}{=} S_e(\phi_e) \, \sim \, \frac{1}{M} \sum_{m=1}^M \mathcal{N}(s|\mu_{e,\theta_m}, \sigma^2_{e,\theta_m}). \label{eq:S_e}
\end{equation}

The networks $f_{\theta_m}$ in (\ref{eq:func_f}) can accept an arbitrary pitch angle $\phi$ to emulate slip distributions $S_e(\phi)$ on edge $e$ for hypothetical angles $\phi$ different from the actual angle $\phi_e$.
This design is useful for subsequent cost derivations. Thus, $f_{\theta_m}$ is implemented in two steps. First, a convolutional neural network (CNN) extracts a feature map from the color map $I$ as $F = \text{CNN}_{\theta_m}(I)$. Subsequently, for each edge $e: v\to v'$, a multi-layer perceptron (MLP) predicts a Gaussian from the feature pair $(F_v, F_{v'})$ and the pitch angle $\phi_e$ as
$[\mu_{e,\theta_m}, \sigma^2_{e,\theta_m}] = \text{MLP}_{\theta_m}(F_v, F_{v'}, \phi_e)$,

The networks $f_{\theta_m}$ are initialized differently and trained individually to minimize the negative log-likelihood loss.
\begin{equation}
L_{\theta_m} = -\frac{1}{|\mathcal{E}|}\sum_{e \in \mathcal{E}} \log\left(\mathcal{N}(s_e^*|\mu_{e,\theta_m}, \sigma^2_{e,\theta_m})\right), \label{eq:loss_func}
\end{equation}
where $s_e^*$ is the ground-truth slip from training maps.

Note that the Gaussian mixture distribution in (\ref{eq:S_e}) has the following forms for the mean and variance:
\begin{align}
\mathbb{E}(S_e) &= \mathbb{E}_{\theta}[\mu_{e,\theta}],\\
\mathbb{V}(S_e) &= \mathbb{E}_{\theta}[\sigma^2_{e,\theta}] + \mathbb{V}_{\theta}[\mu_{e,\theta}], \label{eq:total_variance}
\end{align}
where $\mathbb{E}_{\theta}[X_{\theta}] = \frac{1}{M}\sum_{m=1}^M X_{\theta_m}$ denotes the expectation, and  $\mathbb{V}_{\theta}[X_{\theta}] = \frac{1}{M}\sum_{m=1}^M (X_{\theta_m}-\mathbb{E}_{\theta}[X_{\theta}])^2$ denotes the variance between the ensemble.

We use the ensemble mean $\mathbb{E}(S_e)$ for slip prediction and the total variance $\mathbb{V}(S_e)$ to quantify uncertainty. 
In (\ref{eq:total_variance}), $\mathbb{V}(S_e)$ is decomposed into two components, expressing different types of uncertainties.
The first component, the average of the variance estimates, is known as aleatoric uncertainty and reflects the natural variability 
from the inherent stochasticity of the observations.  
The second component, the variance of the mean estimates, is known as the epistemic uncertainty and reflects the gaps between the training and test data.

In the next section, we derive uncertainty-aware traveling costs using these predictions and omit the subscript $e$ from $S_e$, $\mu_{e}$ and $\sigma^2_{e}$ by focusing on the same edge $e$.

\subsection{Uncertainty-aware Cost Formulation}
\label{section: ucf}
We formulate the cost function $f_{\text{cost}}$ in (\ref{eq:path_planning}) using the DPT predictions provided in the previous section. The edge cost $f_{\text{cost}}$ is based on the predicted travel time, which encompasses both efficiency and safety since faster traverses lead to shorter travel times with lower slips.

The travel time for each edge $e:v \rightarrow v'$ is given by $t_e = ||\boldsymbol{x}(v)-\boldsymbol{x}(v')||/u(e)$, where $\boldsymbol{x}(v)$ denotes the 3D position at node $v$, and $u(e)$ denotes the rover velocity during the edge transition. The velocity $u(e)$ can be calculated from the slip~$s$ using (\ref{eq:slip_ratio}) as follows:
\begin{equation}
    \label{eq:velocity}
    u(e) = 
    \begin{cases}
    \left(1-s\right)u_{\text{ref}},
    &s \geq 0: \text{driving state},\\
    u_{\text{ref}}/\left(1 + s\right),
    &s < 0: \text{braking state}.
    \\
    \end{cases}
\end{equation}

Although treating the predicted travel times directly as costs can reduce the risk of rover immobilization (because $s=1$ results in an infinite travel time), it favors faster rover traverses with larger negative slips, which can endanger the rover. To reduce the risks associated with excessively fast traverse, we adopt the concept of \emph{slip as a risk} introduced in \cite{endo2023risk}. This concept views any slip state other than that on flat ground ($\phi=0$) as a potential risk. Thus, it evaluates the deviations from this stable state as positive values by inverting the sign of the slip $S(\phi)$ for $\phi < 0$ as follows:
\begin{equation}
    S_{\text{risk}}(\phi) =
    \begin{cases}
    S(\phi), 
    &\phi \geq 0: \text{ascent},\\
    2\mathbb{E}[S(0)] - S(\phi),
    &\phi < 0: \text{descent},\\
    \end{cases}
    \label{eq:random_variable}
\end{equation}
where $S(\phi)$ denotes a random slip variable from (\ref{eq:S_e}).

Using the slip as a risk, we formulate an uncertainty-aware slip estimate $\hat{s}_\text{risk}$ as follows:
\begin{subequations}
\label{eq:unc_slip}
\begin{align}
\hat{s}_{\text{risk}}(e) 
& = \mathbb{E}[S_{\text{risk}}(\phi_e)] + \lambda \sqrt{\operatorname{Var}[S_{\text{risk}}(\phi_e)]}, \label{eq:unc_slip_a} \\
& = \mathbb{E}[S_{\text{risk}}(\phi_e)] + \lambda \sqrt{\mathbb{E}_{\theta}[\sigma^2_{\theta}] + \mathbb{V}_{\theta}[\mu_{\theta}]} .\label{eq:unc_slip_b}
\end{align}
\end{subequations}
This estimate computes the expectation of $S_\text{risk}$ using the mixture distribution in (\ref{eq:S_e}) and further pessimistically adjusts this risk estimate by adding the standard deviation of $S_\text{risk}$ scaled by the hyperparameter $\lambda\in \mathbb{R}^+$.

After converting $\hat{s}_{\text{risk}}(e)$ to a risk-aware velocity estimate $\hat{u}_{\text{risk}}(e)$ via the slip-to-velocity conversion in (\ref{eq:velocity}), we finally derive our cost as a risk-aware travel time as follows:
\begin{equation}
    \label{eq:cost_function}
    f_\text{cost}(e:v\to v') = \frac{||\boldsymbol{x}(v) - \boldsymbol{x}(v')||}{\hat{u}_{\text{risk}}(e)}.
\end{equation}
This cost function  evaluates both the traverse efficiency and stability in ascent and descent, while considering the uncertainties in slip predictions.

\subsection{Test-time Domain Adaptation}
\label{section:adaptation}
We incorporate test-time DA into our framework using actual slip experiences observed in-situ during path execution, as illustrated in Fig.~\ref{fig:overview}c.
Specifically, we perform adaptation after each rover movement and re-plan using the updated DPT predictions.
This test-time adaptation is enabled by our end-to-end DPT model $f_{\theta_m}$, which is trained solely from slip experience using the loss function in (\ref{eq:loss_func}), allowing for the backpropagation of the loss computed along the running trajectory.

However, on-the-fly test time adaptation is challenging due to %because of
the limited number of available samples. Without sufficient samples, adapting pre-trained models may unintentionally eliminate valuable knowledge acquired during pre-training.

Inspired by previous work on few-shot test-time adaptation~\cite{zhang2022few}, we propose adapting pre-trained models $f_{\theta_m}$ by adding and tuning a fraction of new parameters while keeping all pre-trained parameters $\theta_m$ fixed. Particularly, after every linear and convolutional layer in $f_{\theta_m}$, denoted by $g(\boldsymbol{x}) = W\boldsymbol{x} + \boldsymbol{b}$, we insert a trainable adaptation layer as %as follows:
\begin{equation}
    \label{eq:adaptation_layer}
    \bar{g}(\boldsymbol{x})=\boldsymbol{\gamma} \odot g(\boldsymbol{x}) + \boldsymbol{\beta},
\end{equation} 
where $\boldsymbol{\gamma}$ and $\boldsymbol{\beta}$ are vectors of new trainable parameters, and $\odot$ is an element-wise multiplication. We initialize $\boldsymbol{\gamma}$ to $\mathbb{1}$ and $\boldsymbol{\beta}$ to $\mathbb{0}$, such that this layer initially preserves the original model behavior, that is, $\bar{g}(\boldsymbol{x}) = g(\boldsymbol{x})$. 

Compared with directly updating $W$ and $\boldsymbol{b}$ in each layer $g$, this adaptation model better preserves the original semantics of each feature channel in $\boldsymbol{y} = g(\boldsymbol{x})$ by avoiding the mixing of $\boldsymbol{y}$ across the channels. Additionally, having significantly fewer parameters in $\boldsymbol{\gamma}$ than in $W$ helps prevent overfitting of the models to a limited number of samples.

\begin{figure*}[t]
    \centering
    \includegraphics[width=1.\hsize]{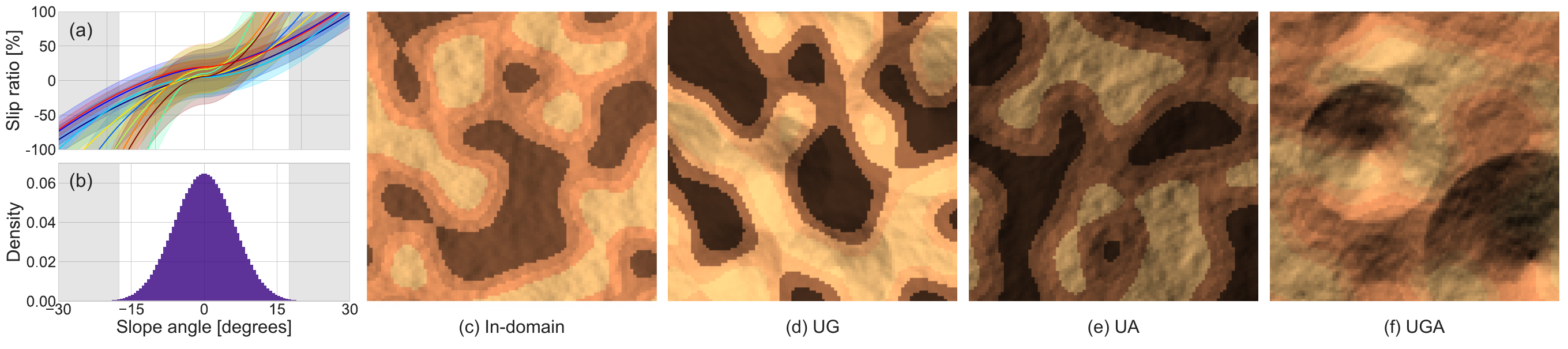}
\caption{
Dataset visualizations. (a) Ten latent slip functions associated with terrain classes, with shaded areas indicating the $2\sigma$ ranges of their additive noises.
(b) Distribution of terrain slope angles in the training subset. The gray shading in (a) and (b) indicates the sampling range of the crater slopes in (d) and (f).
(c)-(f) Example color maps for the in-domain, unfamiliar geometry, unfamiliar appearance, and unfamiliar geometry and appearance test subsets.
}
\label{fig:dataset}
\end{figure*}

\section{Simulation Experiments}

In this section, we validate our rover navigation algorithm through extensive simulations to demonstrate its effectiveness under various uncertainties in traversability prediction.

\subsection{Dataset}
We create a synthetic dataset for path-planning problems, emphasizing the challenge of train--test domain shifts. Thus, we prepare training and validation subsets as in-domain data for training the DPT models and four types of test subsets as out-of-domain data for evaluating navigation performance in unfamiliar environments. 
Each map instance is a 96 $\times$ 96 grid with a resolution of 1 m per grid, providing color and height maps as inputs for navigation and a ground-truth slip map for model training and traverse simulation during testing. The data generation process is as follows:

\subsubsection{Overall data generation}
We define 10 terrain classes in the dataset as hidden information. Each class $c$ is associated with the nonlinear stochastic slip function $f_c(\phi) + \epsilon_c$, where $\epsilon_c \sim \mathcal{N}(0, \sigma_c^2)$, as shown in Fig.~\ref{fig:dataset}a, and appearance color $I_c$.
Subsequently, we create \emph{map templates} by generating random rough terrain geometries using fractal terrain modeling~\cite{yokokohji2004evaluation}.
From each map template, we generate multiple map instances by randomly assigning a terrain class to each node (pixel), along with its associated slip function and color. 
The assignment is performed using random Perlin noise images to form several class clusters on each map.
Slip maps are generated by sampling the slips from the assigned slip functions.
For the color maps, shading is applied to obtain more realistic observation imagery (Fig.~\ref{fig:dataset}c-f). These shaded color maps are computed as $I_v = I_{c(v)} \max(\boldsymbol{n}_v^T \boldsymbol{\ell}, \epsilon)$ for each node $v$, where $\boldsymbol{n}_v$ is a surface normal vector, $\boldsymbol{\ell}$ is a random lighting direction vector, and $\epsilon$ is a small positive constant representing the ambient light level.
For the training, validation, and test subsets,  20,000, 5,000, and 400 map instances are created from 100, 25, and 40 map templates, respectively.

\subsubsection{In-domain data generation}
We control two design factors for data generation to distinguish between in-domain and out-of-domain (OOD) data. The first factor is the steepness of the geometry in the map templates. The other factor is the lighting angle during shading. Specifically, the light direction vector is sampled in spherical coordinates with an azimuth angle $\psi \sim U(0, 2\pi)$ and a vertical component $z \sim U(z_{\text{min}}, z_{\text{max}})$, where $z_{\text{min}}$ and $z_{\text{max}}$ control the elevation angle range.
As in-domain data for training and validation, we adopt fractal terrain geometries, resulting in moderate steepness, with slope angles mostly ranging within $\pm 17^\circ$ (see Fig.~\ref{fig:dataset}b). Additionally, we set the lighting parameters $\{z_{\text{min}}, z_{\text{max}}\}$ to $\{0.8, 1.0\}$.

\subsubsection{Out-of-domain test data generation}
For the OOD geometry setting, we introduce steeper crater-like geometries to the fractal-terrain geometries (Fig.~\ref{fig:dataset}d and f), with their steepness randomly sampled from the range of 17.5$^\circ$--30$^\circ$ (see the shaded regions in Fig.~\ref{fig:dataset}a and b). 
For the OOD appearance setting, we set the lighting parameters $\{z_{\text{min}}, z_{\text{max}}\}$ to $\{0.3, 0.5\}$. By combining these two OOD factors, we create the following four test subsets:
\begin{itemize}
\item \textbf{In-domain subset} (Fig.~\ref{fig:dataset}c) simply follows the same data distributions as the training and validation subsets for evaluating baseline performance under familiar environmental conditions.
\item \textbf{Unfamiliar geometry (UG) subset} (Fig.~\ref{fig:dataset}d) contains  OOD crater-like geometries with steeper slopes.
\item \textbf{Unfamiliar appearance (UA) subset} (Fig.~\ref{fig:dataset}e) contains OOD appearances with darker shading due to lower lighting angles.
\item \textbf{Unfamiliar geometry and appearance (UGA) subset} (Fig.~\ref{fig:dataset}f) combines both OOD factors. 
\end{itemize}
Each test subset contains 100 map instances generated from 10 map templates.
Also, we prepare 40 maps, similar to the test subsets but 10 times smaller in size, for tuning the hyperparameters of planning and adaptation algorithms.

\subsection{Implementation Details}
For DPT, we use $M=10$ networks.
Each network $f_{\theta_m}$ adopts U-Net~\cite{ronneberger2015unet} with ResNet-18~\cite{he2016deep} as the CNN.
We also apply vector embedding to slope angles before feeding them into the MLP.
During pre-training, each network $f_{\theta_m}$ is trained on the training subset for 100 epochs using the Adam optimizer~\cite{kingma2015adam} with a learning rate of $10^{-3}$ and batch size of 256.
The model weights that yielded the lowest validation loss are used for evaluating the test subsets.
Planning is performed using an A* search~\cite{hart1968formal}, which provides an optimal solution for grid maps.
Both planning and execution assume a velocity $u_{\text{ref}}$ of 0.1 m/s.
In each adaptation step, backpropagation runs for 10 iterations with a learning rate of $10^{-3}$ for each $f_{\theta_m}$ to exploit the lowest-loss models for subsequent replanning.

\begin{table*}[t!]
    \centering
    \begin{threeparttable}        
    \caption{Quantitative path planning results on 400 problem instances from the four test subsets}
    \label{tab:quantitative_planning_results}
    \setlength{\tabcolsep}{2.0pt}
    %\scalebox{0.9}{%
        \begin{tabular}{ll|cc|cccccccccccc}
        \toprule
        {} & & \multicolumn{2}{c|}{GT} & \multicolumn{3}{c}{In-domain} & \multicolumn{3}{c}{{UG}} & \multicolumn{3}{c}{{UA}} & \multicolumn{3}{c}{UGA} \\
        %\cmidrule(r){1-2}
        \cmidrule(lr){3-4} \cmidrule(lr){5-7} \cmidrule(lr){8-10} \cmidrule(lr){11-13} \cmidrule(lr){14-16}
        {} & & Slp & Cls & {Suc$\uparrow$} & {$T_{\text{total}}\downarrow$} & {$s_{\text{max}}\downarrow$} & {Suc$\uparrow$} & {$T_{\text{total}}\downarrow$} & {$s_{\text{max}}\downarrow$} & {Suc$\uparrow$} & {$T_{\text{total}}\downarrow$} & {$s_{\text{max}}\downarrow$} & {Suc$\uparrow$} & {$T_{\text{total}}\downarrow$} & {$s_{\text{max}}\downarrow$} \\
        \midrule \midrule
        \multirow{1}{*}{Seg+EV\cite{cunningham2017locally}} & & \checkmark & \checkmark &
        \textbf{99} & {\textbf{24.1}\,$\pm$\,0.9} & 48.5\,$\pm$\,13.3 & 
        93 & {\textbf{24.9}\,$\pm$\,1.5} & 61.1\,$\pm$\,17.6 & 
        84 & \textbf{25.1}\,$\pm$\,1.1 & 64.4\,$\pm$\,19.7 & 
        66 & 29.8\,$\pm$\,34.6 & 77.2\,$\pm$\,20.9 \\
        \multirow{1}{*}{MoE+CVaR\cite{endo2023risk}} & & \checkmark & \checkmark &
        \textbf{99} & 24.4\,$\pm$\,1.1 & \textbf{45.6}\,$\pm$\,12.2 & 
        \textbf{94} & 25.1\,$\pm$\,1.4 & 57.2\,$\pm$\,16.8 & 
        87 & 25.5\,$\pm$\,1.4 & 65.7\,$\pm$\,18.5 & 
        69 & \textbf{26.2}\,$\pm$\,1.9 & 77.6\,$\pm$\,18.8 \\
        \multirow{1}{*}{\textbf{Ours}} & & \checkmark &  &
        {98} & 24.6\,$\pm$\,1.2 & {46.7\,$\pm$\,12.9} & 
        93 & 25.5\,$\pm$\,1.5 & {\textbf{56.4}\,$\pm$\,14.0} & 
        98 & 25.2\,$\pm$\,1.1 & 57.2\,$\pm$\,12.9 & 
        87 & 26.4\,$\pm$\,2.5 & 61.8\,$\pm$\,18.4 \\
        \multirow{1}{*}{\textbf{Ours w/ DA}} & & \checkmark &  &
        98 & 24.6\,$\pm$\,1.3 & 47.9\,$\pm$\,13.3 & 
        91 & 25.4\,$\pm$\,1.6 & 56.6\,$\pm$\,15.8 & 
        \textbf{99} & {25.2\,$\pm$\,1.4} & {\textbf{55.2}\,$\pm$\,13.9} & 
        \textbf{90} & {26.5\,$\pm$\,3.0} & {\textbf{58.6}\,$\pm$\,17.1} \\
        \bottomrule
    \end{tabular}
    %}
    \begin{tablenotes}[flushleft]
    \item[] Suc, $T_{\text{total}}$, and $s_{\text{max}}$ denote success rate [\%], total time [min], and maximum slip [\%], respectively. Suc and $s_{\text{max}}$ indicate path safety, and $T_{\text{total}}$ reflects path efficiency.  $T_{\text{total}}$ and $s_{\text{max}}$ are shown as mean $\pm$ standard deviation. GT indicates the use of ground-truth slip (Slp) and class (Cls) annotations.
    \end{tablenotes}
    \end{threeparttable}
\end{table*}

\subsection{Experimental Setup}
For each instance, the planners search for a path from the start (16 m, 16 m) to the goal (80 m, 80 m). 
After finding a solution, the rover navigates the map while receiving noisy edgewise slips $s_e$. 
A solution is considered successful if the rover never observes $s_e \geq$ 1 (rover immobilization) or $s_e \leq-1$ (dyscontrol situation) along the path toward the goal.
We employ the following three performance metrics:
\begin{itemize}
    \item \textbf{Success rate} (Suc) [\%] denotes the percentage of problem instances with successful solutions out of the total number of instances provided.
    \item \textbf{Total time} ($T_{\text{total}}$) [min] measures the path efficiency as total traversing time per successful path.
    \item \textbf{Maximum slip} (\(s_{\text{max}}\)) [\%] evaluates path safety by calculating \(100 \times \max(\boldsymbol{s_e})\) post-execution, including failure cases.  \(\boldsymbol{s_e}\) denotes the experienced slips; a lower \(s_{\text{max}}\) indicates a safer traverse.
\end{itemize}

We run our method using $\lambda=2.0$ in (\ref{eq:unc_slip}) with and without DA (described in Section~\ref{section:adaptation}).
We compare different prediction models and risk evaluation methods to validate the proposed approach.
As a baseline, we adopt a segmentation-based planner (Seg+EV)~\cite{cunningham2017locally} that selects single class-associated Gaussian processes based on terrain classification and simply calculates the expected values (EVs) of the GPs for planning costs, ignoring prediction uncertainties.
We also evaluate a MoE-based risk-aware planner (MoE+CVaR)~\cite{endo2023risk}, which integrates multiple GPs with terrain classification likelihoods and calculates risk-aware planning costs with CVaR using $\alpha=0.9$.
These methods use the same CNN architecture as ours for terrain classification. However, the training of CNNs and class-associated GPs relies on additional ground-truth (GT) class annotations from the training subset (see GT in Table \ref{tab:quantitative_planning_results}).

\subsection{Results}

\subsubsection{Quantitative Comparison}

Table \ref{tab:quantitative_planning_results} summarizes the path-planning performance across the four test subsets.
Overall, our approach performs comparably to the baseline methods for the in-domain and UG subsets and significantly outperforms them for the UA and UGA subsets, despite using only slip annotations for training.
Comparisons of our results with each baseline and the impact of DA are discussed below.

\emph{vs. Seg+EV}: The Seg+EV-based planner~\cite{cunningham2017locally} tends to prioritize path efficiency over safety because its EV-based planning costs ignore prediction uncertainties. While this tendency results in the lowest $T_\text{total}$ for all subsets but UGA, our approach maintains a comparably low $T_\text{total}$ with much higher safety for the GA and UGA subsets.
These benefits stem from our time-based cost evaluation using uncertainty-aware slip estimation in (\ref{eq:unc_slip}), which balances traversability and travel time and is beneficial for successful planetary explorations that require both safety and efficiency.

\emph{vs. MoE+CVaR}: Although the MoE+CVaR-based planner~\cite{endo2023risk} also incorporates uncertainties in slip predictions for risk-aware planning, it experiences low success rates for the UA and the most challenging UGA subsets. 
We found that the darker shaded regions in these subsets often cause misclassifications by the terrain classifier in the MoE slip model.
Conversely, our DPT fuses color- and geometry-based features to predict slip distributions end to end, effectively disambiguating terrain colors and shading effects to improve predictions.

\begin{table}[t]
    \centering
    \caption{Domain adaptation effects for slip prediction}
    \label{tab:quantitative_prediction_results}
    \begin{threeparttable}
    \begin{tabular}{l|cccc}
        \toprule
         & In-domain & {UG} & {UA} & UGA \\
        \midrule \midrule
        {Before DA} & \textbf{8.7} $\pm$ 1.6 & 12.3 $\pm$ 2.6 & 12.6 $\pm$ 1.2 & 16.3 $\pm$ 2.3 \\
        \textbf{After DA} & \textbf{8.7} $\pm$ 1.5 & \textbf{11.7} $\pm$ 1.8 & \textbf{11.1} $\pm$ 1.3 & \textbf{14.3} $\pm$ 2.1 \\
        \bottomrule
    \end{tabular}
    \begin{tablenotes}[flushleft]
    \item[] The 100 times scaled mean absolute errors of the slip predictions by the proposed DPT model before and after the domain adaptation.
    \end{tablenotes}
    \end{threeparttable}
\end{table}

\emph{Impact of DA}: DA successfully improves the success rates and $s_\text{max}$ scores for the UA and the most challenging UGA subsets.
For the in-domain subset, the DA maintains the same performance as that before adaptation, as the new in-situ samples fall within the pretraining domain.
However, for the UG subset, we observe a slight decrease in the success rate.
This subset poses the difficulty of adapting and correctly extrapolating the learned slip functions to OOD slope angles (\ie, $|\phi| > 17^\circ$) given a few OOD samples.
Because the current architecture of our slip model does not strictly follow the monotonicity inherent in latent slip functions $s = f(\phi)$, such an extrapolation through few-shot adaptation may be challenging.
Incorporating monotonicity into the DPT slip model may improve its performance and adaptation stability.
Nevertheless, we found that DA consistently improves the average slip prediction accuracy across the four test subsets.
The results are provided in Table~\ref{tab:quantitative_prediction_results}, comparing the mean absolute errors of the slip predictions for entire map regions before and after test-time adaptation.

\begin{figure*}[t]
    \centering
    \includegraphics[width=1.\hsize]{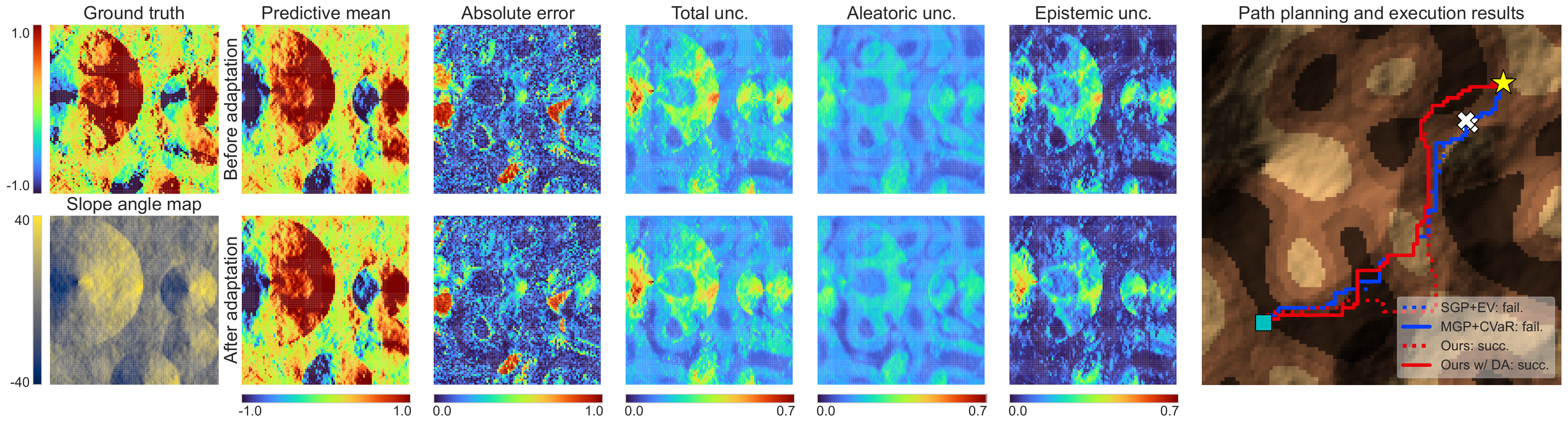}
    \caption{
Traversability predictions and navigation results for a map in the unfamiliar geometry and appearance subset. 
The left column shows the ground-truth slips (top) and slope angles (bottom) for the rightward movements. 
The middle columns display slip predictions before (top row) and after (bottom row) domain adaptation, showing absolute errors with total, aleatoric, and epistemic uncertainties.
The right column shows path planning and execution results. 
The cyan square and yellow star indicate the start and goal locations, respectively; white crosses mark locations where rovers failed to traverse.}
    \label{fig:qualitative_results}
\end{figure*}

\subsubsection{Qualitative Discussion}

Fig.~\ref{fig:qualitative_results} shows the probabilistic slip predictions for rightward movements in a UGA subset problem instance, comparing the results before and after DA.

In the results of the pre-trained DPT model before adaptation (top row), the absolute error map shows robust slip predictions, except in OOD terrain areas, such as crater-type terrain with steep inclinations and intense shading. The errors in these areas are highly correlated with the intense uncertainties in the total uncertainty map.
This capability provides a comprehensive assessment of the predictions and their confidence, helping to identify both hazardous and unfamiliar terrain conditions.
Uncertainty decomposition further reveals epistemic components around OOD areas and aleatoric components aligned with terrain class patterns owing to class-dependent noise in the latent slip functions.

In the results of the adapted DPT model (bottom row), we observed reduced errors for OOD regions while maintaining the overall prediction trends. This robust adaptation is enabled by our adaptation model, which tunes only the scale and bias parameters.

On the right side of Fig.~\ref{fig:qualitative_results}, we compare the planned paths obtained using our method and baseline methods. Our method achieves more robust path planning than the other methods by integrating these well-calibrated uncertainty estimates into traversability evaluation.
The MoE+CVaR-based planner also exploits this uncertainty. However, its misallocated class probabilities cause the same traverse failure as that caused by the Seg+EV-based planner.

\subsubsection{Parameter Study}

\begin{figure}[t]
    \centering
    \includegraphics[width=1.\hsize]{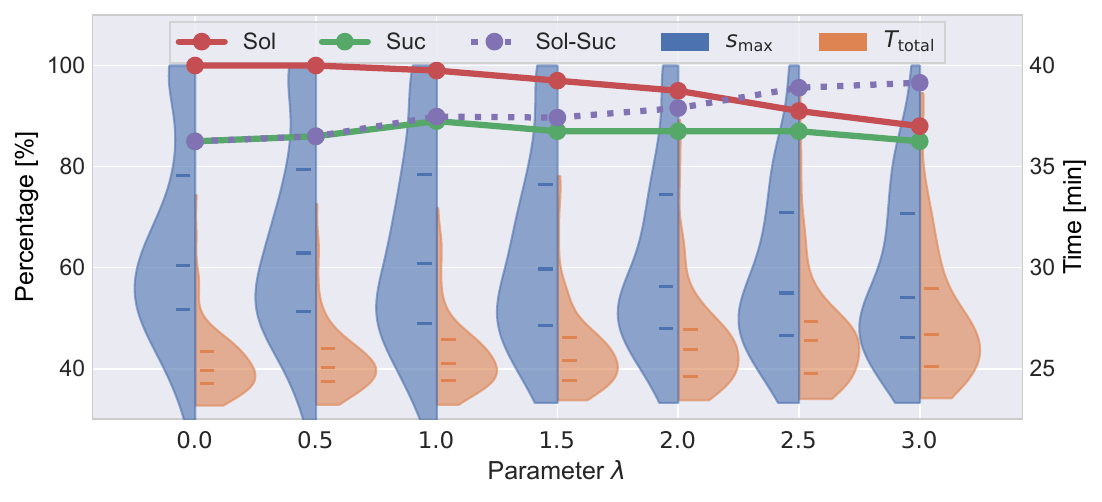}
    \caption{
Parameter study for the unfamiliar geometry and appearance subset, showing the variations of five metrics with varying $\lambda$.
The left percentile y-axis represents Sol, Suc, Sol-Suc, and $s_{\text{max}}$, while the right y-axis represents  $T_{\text{total}}$.
Blue and orange violin plots represent the distribution of $s_{\text{max}}$ and $T_{\text{total}}$, respectively, with lines marking the 25th, 50th, and 75th percentiles.}
\label{fig:parameter_study_results}
\end{figure}

We assess the impact of uncertainty exploitation for the UGA subset, by running our algorithm using $\lambda \in \{0.0, 0.5, \ldots, 3.0\}$ without DA.
We introduce additional performance metrics: \emph{solved rate} (Sol) [\%] as the percentage of problem instances for which a planner finds solutions and \emph{solved success rate} (Sol-Suc) [\%] as the success rate within solved cases.
Fig. \ref{fig:parameter_study_results} illustrates how $\lambda$ controls the balance between safety and efficiency: higher $\lambda$ values guide rovers to less slippery paths, while lower values allow more aggressive maneuvers to reduce travel time.
For safer navigation, higher $\lambda$ is preferable. Although increasing $\lambda$ results in lower solution rates, it improves solved success rates, suggesting that planning failures effectively predict serious immobilization risks in the environment.
The optimal $\lambda$ depends on the mission specifics and environment, but statistical considerations offer general guidelines. 
Accounting for prediction errors within the 2$\sigma$ range (95\% confidence interval) using $\lambda=2$ balances the safety and efficiency across various scenarios.

\subsubsection{Open Issues and Possible Extensions}
\label{section:open_issues_possible_extensions}

Further refinement of traversability modeling and uncertainty exploitation can improve risk-aware planning.
Potential improvements include the adoption of monotonic neural networks for slip modeling~\cite{kim2024scalable} and statistical theories, such as CVaR, for risk assessment. 
We can also explore active learning strategies~\cite{endo2022active} using our framework by leveraging the epistemic uncertainty components of the DPT as cues for regions that require more training samples. However, ensuring the safety of rovers during such risk-taking behaviors remains critical.

\section{Conclusion}

We have introduced a unified learning and planning framework for navigating rovers on deformable terrain in celestial environments, with a particular focus on ML-based traversability prediction and uncertainty handling.
Extensive simulation experiments have demonstrated that our approach enables safe rover traverse with improved success rates, even under unfamiliar geometries and appearances of terrains.
Test-time DA through the end-to-end architecture of our ML model also contributes to successful navigation by continuously reducing prediction errors as the rovers traverse.
Future work will explore an active learning strategy to reduce uncertainties in the environment efficiently through DA.
Additionally, we plan to investigate the feasibility of our method using real-world data, such as the Martian terrain data ~\cite{ono2016data}.

%                                   % the textheight of the last page by a suitable amount.
%                                   % This command does not take effect until the next page
%                                   % so it should come on the page before the last. Make
%                                   % sure that you do not shorten the textheight too much.

% \section*{acknowledgement}
% This work was partially supported by JSPS KAKENHI Grant Number JP22J22731. The authors thank Ryo Yonetani during the early stages of this study.

\bibliographystyle{IEEEtran}
\bibliography{IEEEexample}

\end{document}